\documentclass{article}
\usepackage{ijcai23}

\usepackage{times}
\usepackage{soul}
\usepackage{url}
\usepackage[hidelinks]{hyperref}
\usepackage[utf8]{inputenc}
\usepackage[small]{caption}
\usepackage{graphicx}
\usepackage{amsmath}
\usepackage{amsthm}
\usepackage{booktabs}
\usepackage{algorithm}
\usepackage{algorithmic}
\usepackage[switch]{lineno}

\usepackage{multirow}
\usepackage{makecell}
\usepackage{array}
\usepackage{graphicx}
\usepackage{float}
\usepackage{caption}

\usepackage{subfigure}
\usepackage{arydshln}

\usepackage{CJKutf8}

\title{Chinese Cyberbullying Detection: Dataset, Method, and Validation}

\author{
Yi Zhu$^1$$^,$$^2$\and
Xin Zou$^1$\and
Xindong Wu$^2$$^,$$^3$
\affiliations
$^1$School of Information Engineering, Yangzhou University, Yangzhou {\rm 225009}, China\\
$^2$Key Laboratory of Knowledge Engineering with Big Data (Hefei University of Technology), Ministry of Education, Hefei {\rm 230009}, China\\
$^3$School of Computer Science and Information Engineering, Hefei University of Technology, Hefei {\rm 230009}, China
\emails
zhuyi@yzu.edu.cn,
mz120241002@stu.yzu.edu.cn,
 xwu@hfut.edu.cn
}

\begin{document}
\begin{CJK}{UTF8}{gbsn}

\maketitle

\begin{abstract}

Existing cyberbullying detection benchmarks were organized by the polarity of speech, such as "offensive" and "non-offensive", which were essentially hate speech detection. However, in the real world, cyberbullying often attracted widespread social attention through incidents. To address this problem, we propose a novel annotation method to construct a cyberbullying dataset that organized by incidents. The constructed CHNCI is the first Chinese cyberbullying incident detection dataset, which consists of 415,463 comments in 195 incidents. Specifically, we first combine three cyberbullying detection methods based on explanations generation as an ensemble method to generate the pseudo labels, and then let human annotators judge these labels. Then we propose the evaluation criteria for validating whether it constitutes a cyberbullying incident. Experimental results demonstrate that the constructed dataset can be a benchmark for the tasks of cyberbullying detection and incident prediction. To the best of our knowledge, this is the first study for the Chinese cyberbullying incident detection task.

\end{abstract}

\section{Introduction}

Cyberbullying has become a pervasive issue and attracted widespread social attention, which manifests in diverse online platforms and often causing severe emotional, psychological, and societal repercussions \cite{mahmud2023cyberbullying,dos2024violation}. In recent decades, cyberbullying detection aims to identify and mitigate abusive content promptly, thereby safeguarding online environments and ensuring user well-being \cite{bozyiugit2021cyberbullying}. Despite extensive research conducted on cyberbullying detection in various languages, including English \cite{kim2021human,salawu2017approaches}, German \cite{fischer2020traditional,schultze2013emotional},  Russian \cite{boronenko2013topicality,kintonova2021cyberbullying}, and Arabic \cite{albayari2021cyberbullying,musleh2024machine}, Chinese cyberbullying detection has received limited attention. In this paper, we address this gap by focusing on the Chinese cyberbullying detection task.

To enable the development and evaluation of effective Chinese cyberbullying detection methods, a large-scale dataset is intuitively important. Existing widely used English cyberbullying benchmarks, such as Cyberbullying\underline{ }Tweets \cite{wang2020sosnet}, Formspring \cite{reynolds2011using}, and MySpace \cite{kumar2022bi}, have provided valuable resources for developing and evaluating models aimed at detecting cyberbullying and related abusive behaviors online. However, these datasets were organized by the polarity of speech, which have the following two problems.

(1) Low Coverage: The datasets classified by the polarity of speech may have limitations in addressing real practical problems, and these datasets are actually also applied for hate speech detection. However, just judging comments as "cyberbullying" and "non-cyberbullying" or "offensive" and "non-offensive" is far from enough in real-world cyberbullying related tasks, such classification does not capture the temporal dynamics or social amplification of cyberbullying incidents, which often escalate rapidly and cause widespread harm before interventions can be deployed. For example, a single offensive comment may not be problematic in isolation, but when thousands of similar comments appear in a short time frame targeting a specific individual or group, the cumulative effect can be severely damaging. 
%how to avoid the harm caused by large-scale cyberbullying is ignored. For example, cyberbullying would be much less harmful if incidents could be predicted before a large-scale outbreak rather than detected after.

(2) High Cost: Annotating the categories of sentences is a time-consuming and labor-intensive task. It requires human annotators to carefully consider suitable categories, taking into account the humanistic histories or cultural stories behind the text. For example, given the sentence as "The famous German landmark in 1941: The Louvre.", the true intention behind the statement is "Louvre is French, in 1941, the German Nazi regime occupied France in the World War II.", so the sentence is indeed an offensive comment. Due to the complexity of the task, annotating a large number of instances becomes challenging within a reasonable timeframe and budget.

To address these challenges, we propose a novel annotation method to construct a cyberbullying dataset that organized by incidents. Firstly, we propose an ensemble method that leverages three cyberbullying detection methods based on explanations generation to generate the pseudo labels. This automated methods can quickly generate potential labels and corresponding explanations, reducing the burden on human annotators. Then we let human annotators assess these pseudo labels, this collaborative process harnesses the expertise of human annotators while leveraging the efficiency and scalability of machine-generated labels. Secondly, we propose the evaluation criteria for judging whether it constitutes a cyberbullying incident, and the following incident prediction can be validated. All these efficiencies allow for the creation of a larger dataset within a reasonable budget.

Our work is motivated by the following two findings:

(1) Cyberbullying incidents are more specific and impactful than general cyberbullying behaviors. While general cyberbullying involves isolated abusive interactions, cyberbullying incidents refer to sustained, event-related abusive behaviors that escalate over a short period, often leading to serious consequences and widespread social attention. These cyberbullying incidents amplify harm due to their concentrated nature and the scale of participation, making their detection and prediction crucial for timely intervention. Notably, in the experiments, the collected incidents are the real-world events or trending topics that gain widespread public attention on Chinese social media. Cyberbullying incidents are identified within these events; however, not all such events necessarily involve cyberbullying behavior.

(2) Machine-generated cyberbullying detection methods based on explanations can introduce interpretability for detection results. By leveraging detection techniques like language models, paraphrasing models, or multi-agents, the explanations for why it is detected as cyberbullying can be generated. These explanations enrich the dataset by identifying the humanistic histories and cultural stories behind the sentences, capturing the true intentions for cyberbullying detection. Assessing the rationality of these explanations is much simpler for the annotator compared to making decisions without prior knowledge.

In summary, our contributions are listed below:

(1) We propose a novel annotation approach based on human and machine collaboration to construct a cyberbullying dataset that organized by incidents. Our approach provides a good idea for constructing large-scale, high-coverage datasets, especially for tasks involving distinguishing cyberbullying that are characterized by unclear decision boundaries. Based on our method, we construct the first large-scale Chinese Cyberbullying Incident dataset CHNCI that consists of 415,463 comments in 195 incidents, which cover five different text genres, namely Business, Entertainment, Sports, Society and Politics.

(2) We present three cyberbullying detection methods based on explanations (paraphraser-based, Chain-of-Thought (CoT)-based, and multi-agents-based), and give an ensemble method that combines the three methods. Experimental results on CHNCI show that the ensemble method can be served as a strong baseline for future studies.

(3) We provide evaluation criteria for determining whether the statements surrounding a certain event would constitute a cyberbullying incident, then the tasks of incident prediction can be further validated. Experimental results demonstrate that the constructed CHNCI can effectively reflect the development trend of cyberbullying incidents.

The dataset and code are available at https://github.com/zhuyiYZU/CHNCI.

\section{Related Work}
\subsection{Cyberbullying Detection Resources}

Existing cyberbullying detection datasets are available for various languages, including English and other languages. In existing datasets, each instance is composed of a sentence, some attributes, and corresponding labels.

In English, the first cyberbullying detection dataset from Formspring \cite{reynolds2011using}, which has been subject to updates throughout the years. When Formspring was first created in 2011, it had nearly 4000 samples, but it has since tripled in size to 2018 \cite{rosa2018deeper}. The ratio of cyberbullying instances it contains is 0.194, which refers to about 2,500 bullying text. By and large, even in English, there are very few standard datasets available for cyberbullying detection \cite{rosa2019automatic}. Although most studies recur to the same social networks in order to obtain data (e.g., Twitter, YouTube), the datasets are independently created by using a publicly available API or scrapping the website for samples. For example, cyberbullying\underline{ }Tweets dataset \cite{wang2020sosnet} comes from Twitter, comprises over 47,000 tweets annotated with cyberbullying labels, including non-cyberbullying comments and those categorized into five distinct bullying types: Age, Ethnicity, Gender, Religion, and Other. To maintain data balance, approximately 8,000 records are allocated to each category. Recently, the repeating datasets among different works have been from MySpace \cite{kumar2022bi}, which is one of the largest datasets used for cyberbullying detection, containing over 381,000 posts organized into approximately 16,000 threads. The dataset captures conversations from the MySpace platform, with posts authored by a diverse demographic, including 34\% female and 64\% male contributors. Each thread represents an interactive discussion, providing a rich source of contextual information for analyzing cyberbullying behaviors. The dataset is annotated with labels identifying instances of cyberbullying, offering a valuable resource for studying patterns, linguistic cues, and user interactions associated with cyberbullying. Due to its size and detailed annotations, the MySpace dataset is often employed to develop and evaluate models for cyberbullying detection and to explore gender-based differences in abusive online behavior.

Besides English, the German cyberbullying detection dataset originated from the GermEval 2018 and contains around 8,000 instances collected from Twitter, with a proportion of approximately 0.34 instances of cyberbullying. The Russian dataset 2019 is the collection of annotated comments from Russian online communication platforms, which consists of a total of 14,412 comments, the ratio of cyberbullying instances is about 0.334. The annotations were validated with the help of Russian language speakers (laypeople) using a crowd sourcing application. All the above datasets in all languages are constructed by human annotators and organized by the polarity of text. Due to their relatively small size, all of these datasets can only be used for evaluation and not for training. 

Unfortunately, research on Chinese cyberbullying detection is still scarce, although there are already some existing datasets, they are either not specifically designed for Chinese cyberbullying detection \cite{deng2022cold} or are not publicly available and accessible \cite{yang2025sccd}. More importantly, there is currently no publicly available cyberbullying datasets organized by incidents, nor any benchmark dataset suitable for evaluating the performance of cyberbullying detection models in incident-based scenarios.
% Wikipedia Talk Labels \cite{wulczyn2017ex}, OLID (Offensive Language Identification Dataset) \cite{zampieri2019predicting}, and Cyberbullying\underline{ }Tweets \cite{wang2020sosnet}, 

\subsection{Cyberbullying Detection Methods}

Based on the different techniques to feature learning, traditional cyberbullying detection methods can roughly be categorized into four groups: machine learning-based methods \cite{balakrisnan2023cyberbullying}, dictionary-based methods \cite{mahbub2021detection}, rule-based methods \cite{chong2022comparing}, and hybrid methods \cite{raj2021cyberbullying}. (1) Machine learning-based methods typically employ classification models such as Support Vector Machines (SVM) for detecting cyberbullying. For example, Raisi et al. proposed a weakly supervised machine learning approach that relies on a limited set of bullying-related vocabulary provided by experts \cite{raisi2017cyberbullying}, which automatically infers users' bullying tendencies and language through the analysis of extensive social media interaction data. (2) Dictionary-based detection methods use predefined word lists to identify cyberbullying. For example, Wang et al. introduced a cyberbullying detection algorithm based on FastText and word similarity metrics \cite{wang2020cyberbullying}. They constructed a list of cyberbullying-related words from the dataset and assessed whether a text contained cyberviolence risks based on text similarity. (3) Rule-based matching methods involve applying predefined rules to match texts with bullying behaviors. For example, Chong et al. compare zero-shot text classification and rule-based matching approaches for identifying cyberbullying behaviors on social media, demonstrating that zero-shot models, particularly using BART-based architectures, outperform rule-based methods in recognizing behaviors like flaming, though both methods struggle with broader behavior detection accuracy \cite{chong2022comparing}. (4) Hybrid frameworks combine multiple methods based on the characteristics of the data to achieve better classification results. For instance, Almomani et al. proposes a hybrid framework combining deep learning and traditional machine learning to detect cyberbullying in images on social media platforms \cite{almomani2024image}. By utilizing pre-trained CNN models (e.g., InceptionV3, ResNet50, VGG16) as feature extractors and feeding these features into classifiers such as Logistic Regression and SVM, the approach achieves improved detection accuracy.

Recently, deep learning methods have been increasingly employed in cyberbullying detection, owing to their ability to learn abstract features by disentangling underlying explanatory factors in the data. For instance, Iwendi et al. conducted an empirical study to evaluate the performance of traditional deep learning models, including LSTM, Bidirectional LSTM (BiLSTM), and RNN \cite{iwendi2023cyberbullying}. The study involved applying data pre-processing steps, and the results highlighted the effectiveness of these deep learning approaches. Additionally, Maity et al. proposed a graph neural network-based multitask framework for sentiment-aided cyberbullying detection \cite{maity2022mtbullygnn}. The GNN framework accurately identifies unlabeled or noisily labeled nodes (sentences) by aggregating information from similarly labeled nodes.

Despite the significant advantages of these deep methods, they are still limited by the effects of labeled data. More recently, with the widespread use of pre-trained language (PLMs) and Large Language Models (LLMs) \cite{qiang2023natural}, these methods have shown to be extremely helpful in cyberbullying detection. For example, Yadav et al. proposed to a PLMs-based cyberbullying detection method \cite{yadav2020cyberbullying}, which introduced pre-trained BERT model with a single linear neural network layer on top as a classifier. Kaddoura et al. evaluated the efficacy of open-source LLMs, e.g., Mistral 7B and Llama3, against the transformer-based model on cyberbullying detection \cite{kaddoura2025language}.

\section{Creating CHNCI}

\begin{figure*}[t]
	\centering
	\includegraphics[width=\linewidth]{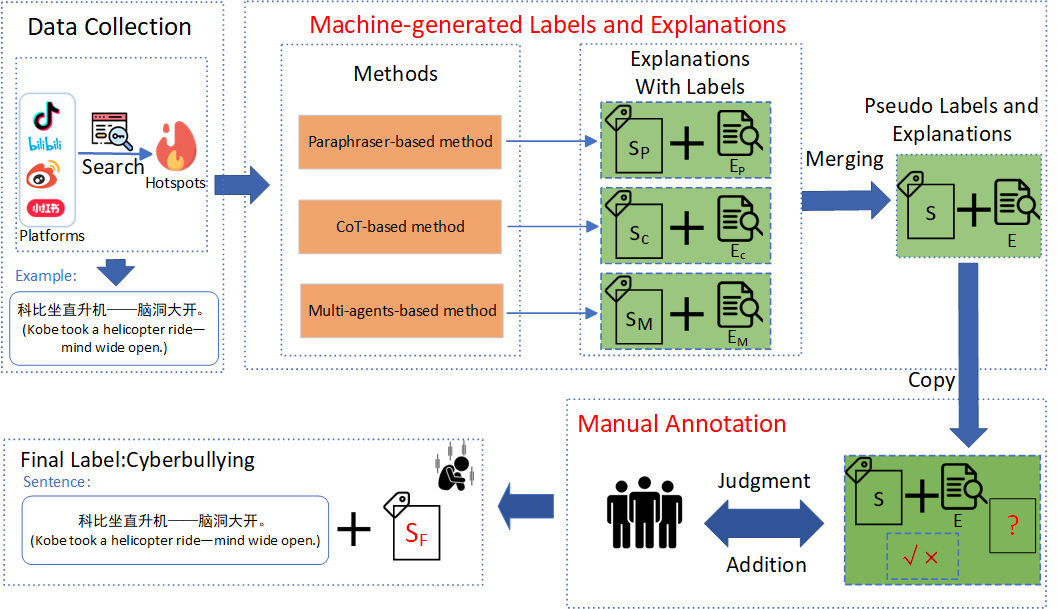}
	\caption{
		The overview of our method for building Chinese cyberbullying detection dataset organized by incidents. The data are collected from multiple mainstream Chinese social media platforms. Our method is composed of two phrases: machine-generated pseudo labels and manual annotation. The first phase combines three cyberbullying detection methods based on explanations as an ensemble method to generate the pseudo labels. The second phase utilizes native Chinese annotators to judge the pseudo labels with generated explanations.
	}
	\label{framework}
\end{figure*}

In this section, we describe our method to build a cyberbullying detection dataset organized by incidents, and the overall architecture for constructing this Chinese dataset is illustrated in Figure \ref{framework}.

\subsection{Data Preparation}

In this step, we first extract the text for data preparation. To ensure diversity and complexity in our dataset, we utilize five distinct text genres: Business, Entertainment, Sports, Society and Politics. Here, genre does not refer to the linguistic style or literary form of the cyberbullying comments themselves, but rather to the topical category or domain of the incident that the comments revolve around. That is, the genre is based on the nature of the event that gave rise to the online discussion. For example, if the cyberbullying occurs in response to the "Tesla Owner Rights Defense Incident," the genre is categorized as "Business". This genre-based organization allows us to analyze how cyberbullying behavior varies across different real-world contexts. By incorporating multiple categories, we aim to capture the richness and intricacy of the Chinese language. 

During the data preparation phase, we extracted relevant information about incidents, including the content of each online comment, its timestamp, and the platform of origin. The incidents here refer to the real-world events or trending topics that gain widespread public attention on Chinese social media. Considering the rapid spread and fermentation of public opinion surrounding incidents, we selected multiple mainstream Chinese social media platforms as data sources, including Douyin\footnote{\url{https://www.douyin.com/}}, Weibo\footnote{\url{https://weibo.com/}}, Xiaohongshu\footnote{\url{https://www.xiaohongshu.com/}}, and Bilibili\footnote{\url{https://www.bilibili.com/}}.
These platforms represent the primary arenas of Chinese social media, encompassing a broad user base and diverse forms of discussion. To discover such incidents, we monitored the top-trending topics on these platforms, which provide event-based or topic-tagged aggregations of posts and discussions. For an identified incident, the relevant contents of user comments and posts are directly collected from the associated topic pages or event tags, ensuring that all retrieved content was contextually aligned with the selected incident. This method guarantees that the messages in our dataset are highly relevant to the real-world event being studied.

By integrating data from multiple platforms, we aimed to comprehensively capture the diversity, complexity, and rapid dissemination characteristics of cyberbullying incidents. Using a custom-designed web crawler, we successfully collected online data for nearly 200 incidents, with about 2,100 comments gathered per incident from various platforms.%To extract representative samples from the extensive gathered data, we employed a stratified sampling method. Given the varying total number of comments for each incident, we first determined an appropriate sampling interval based on the total comment count, then uniformly selected one data point from each interval, ensuring the balance and chronological integrity of the samples.

\subsection{Machine-generated Pseudo Labels}

Considering the sentence $x$, we employ cyberbullying detection methods based on explanations to generate the pseudo labels and corresponding explanations. For a more accurate detection result, we adopt an ensemble approach that combines three distinct methods: paraphraser-based, CoT-based, and multi-agents-based method. By leveraging these diverse methods, each of which taps into different semantic knowledge, we aim to enhance the overall diversity of explanations available for consideration. 

\textbf{Pseudo Labels Generation.} We present three baseline approaches by adapting existing cyberbullying detection methods based on explanations:

(1) Paraphraser-based: The paraphraser-based method utilized the open-source LLMs to paraphrase the given input to generate explanation and pseudo label. The predictions are made through a conversational approach as "Please provide an explanation of the text and predict whether it is cyberbullying". In the experiments, the Llama3 \cite{dubey2024llama} is introduced as LLM and the text is input with a 5-shot prompt.	

(2) CoT-based: The CoT-based method \cite{huang2023chain} employs a Chain-of-Thought prompt to generate high-quality explanations for cyberbullying detection. The CoT prompt templates used in this study are shown in Table \ref{tab:cot_prompts}, which are designed to guide the model through a step-by-step reasoning process to evaluate online comments for cyberbullying detection. Each template directs the model to analyze different aspects of the comment, including offensive language, targeting of individuals or groups, tone, emotional impact, and the broader context of trending events. By breaking down the analysis into distinct steps, the model is able to generate more accurate and detailed explanations for its decisions.

\newcommand{\tabincell}[2]{\begin{tabular}{@{}#1@{}}#2\end{tabular}}  %??????
\begin{table*}[t]
	\centering
	\caption{Chain-of-Thought Prompt Templates for Cyberbullying Detection}
	\label{tab:cot_prompts}
	\renewcommand{\arraystretch}{1.2} % ????
	\small
	\resizebox{\textwidth}{!}{ % ??????????
		\begin{tabular}{|c|p{0.85\textwidth}|} % ????????
			\hline
			\textbf{No.} & \textbf{Prompt Template} \\
			\hline
			1 & \textit{Please analyze the following comment step-by-step. First, identify if there are any offensive or harmful words. Second, determine if the comment targets an individual or group. Finally, decide if it constitutes cyberbullying.} \\
			\hdashline
			2 & \textit{Evaluate the following online comment. Start by examining the tone and wording. Then, assess the intent and potential emotional impact on the recipient. Conclude whether the comment is cyberbullying.} \\
			\hdashline
			3 & \textit{This comment was posted in the context of a trending event. Please break down the statement, identify any signs of aggression or personal attack, and make a final judgment on whether it's cyberbullying.} \\
			\hdashline
			4 & \textit{Given the text below, follow these steps: (1) Look for negative or abusive language (2) Determine if there is a target (3) Assess the severity and impact (4) Provide a decision on cyberbullying.} \\
			\hdashline
			5 & \textit{Analyze the user's comment carefully. Begin by detecting any offensive language. Then check whether it's directed at someone. Explain your reasoning before classifying the comment as bullying or not.} \\
			\hline
		\end{tabular}
	}
\end{table*}
%(3) Multi-agents-based: The multi-agents-based method \cite{guo2024large} generated the explanations and introduce a double-layer multi-agent voting strategy for cyberbullying detection.

%Given multiple prompts as multi-agents, a rule requiring the LLMs to generate the explanations for each input is established, after operation of the $i^{th}$ agent with prompt $p_i$, each output includes both an "Explanation" and a "Label". Since the results of the $i^{th}$ agent have been performed for $in$ times, the internal voting is conducted, it refers to that if the results indicating $y=1$ (cyberbullying) outnumber those indicating $y=0$ (non-cyberbullying), the final decision of the $i^{th}$ agent is classified as "cyberbullying". Finally, the decisions of all agents are voted, i.e., the external voting is conducted among all agents with the majority-voting principle to finalize the detection.
(3) Multi-agents-based: The multi-agent-based method generates explanations and introduces a two-layer multi-agent voting strategy for cyberbullying detection. In this approach, a multi-agent system is employed where multiple independent agents collaborate to process the task. Each agent operates based on a distinct prompt template, enhancing the model's robustness and accuracy.

Specifically, five different agents are designed, each corresponding to a prompt templates that generates a decision for each comment, the details of these agents are presented in Table \ref{tab:prompt_templates}. The voting among these agents constitutes the external voting. Each agent processes the input using its designated prompt template and produces both an "Explanation" and a "Label". The final decision is determined by majority voting across all five agents. If the majority of agents classify the comment as cyberbullying ($y = 1$), the comment is labeled as cyberbullying. Additionally, each agent performs an internal voting process. That is, each agent executes its prompt three times for the same input and votes based on the results. The purpose of internal voting is to reduce the potential bias of a single execution, ensuring the agent's final decision is more robust. If the number of cyberbullying predictions ($y = 1$) exceeds the number of non-cyberbullying predictions ($y = 0$) in the three runs, the agent's final decision is set to cyberbullying.

\begin{table*}[htbp]
	\centering
	\caption{Five Agents Used in the Multi-Agents-based Approach}
	\label{tab:prompt_templates}
	\renewcommand{\arraystretch}{1.2}
	\resizebox{\textwidth}{!}{ % ?????????????????
		\begin{tabular}{|c|c|}
			\hline
			\textbf{Agent} & \textbf{Prompt Template} \\
			\hline
			\textbf{Agent 1} & \tabincell{c}{Read the following comment and determine whether it contains any form of cyberbullying.\\ Provide your reasoning and output a final label: "Cyberbullying" or "Non-Cyberbullying".} \\
			\hline
			\textbf{Agent 2} & \tabincell{c}{Analyze the text below and decide if it exhibits cyberbullying behavior. Explain your\\ reasoning and give a label ("1" for cyberbullying, "0" for non-cyberbullying).} \\
			\hline
			\textbf{Agent 3} & \tabincell{c}{You are an expert in online safety. Review the comment and judge whether it should be\\ classified as cyberbullying. Justify your answer and output the classification result.} \\
			\hline
			\textbf{Agent 4} & \tabincell{c}{Evaluate the given comment and assess whether it constitutes cyberbullying. Include your\\ reasoning process and conclude with a label: cyberbullying or not.}\\
			\hline
			\textbf{Agent 5} & \tabincell{c}{Determine if the following comment is an instance of cyberbullying. Write a brief\\ explanation and assign a label (Cyberbullying/Non-Cyberbullying).} \\
			\hline
		\end{tabular}
	}
\end{table*}

Through this double-layer voting strategy, the model first stabilizes each agent's output through internal voting and then aggregates all agents' final decisions via external voting. This method significantly improves the accuracy and reliability of cyberbullying detection.

\textbf{An ensemble Method.} Considering that each classification method produces different detection results, and to avoid overwhelming annotators with excessive workload or causing fatigue that might affect label quality, we decided to combine the results from these three methods. Specifically, we assigned voting weights of 1 to paraphraser-based, CoT-based, and multi-agents-based methods individually. Using these three approaches, we selected all comments labeled as either "cyberbullying" and "non-cyberbullying" as pseudo-labels. This selection process ensures that the labels generated by multiple methods are more likely to reflect potential classification outcomes, enhancing the reliability and diversity of the labels.

\subsection{Manual Annotation}

To ensure the accuracy of the annotations, we engage multiple annotators for annotation. Specifically, the annotation team consisted of several graduate students majoring in linguistics and a social media practitioner with extensive experience in platform management, all of whom are long-term active users of major Chinese social media platforms. These annotators have maintained their social media accounts for more than five years and spend over 20 hours per week engaging in online activities, which enables them to develop a deep understanding of offensive expressions and their contextual usage in online environments.

\begin{table*}[htbp]
	\centering
	\small % ??????
	\caption{Demographic and Activity Profile of Annotators}
	\label{tab:annotator_profile}
	\renewcommand{\arraystretch}{1.2}
	\resizebox{\textwidth}{!}{ % ?????????????????
		\begin{tabular}{cccccc}
			\toprule
			\textbf{Total Annotators} & \textbf{Gender (M/F)} & \textbf{Age Range} & \textbf{Avg. Registration Time} & \textbf{Avg. Weekly Active Hours} \\
			\midrule
			3 & 2 / 1 & 22 -- 35 yrs & \textgreater 5 years  & \textgreater 20 hours \\
			\bottomrule
		\end{tabular}
	}
\end{table*}

Before starting the annotation process, we provided them with a detailed definition of cyberbullying to ensure they had a clear and comprehensive understanding of the annotation task. In the context of this study, cyberbullying refers to language or behavior in online platforms that intentionally insults, threatens, humiliates, or maligns individuals or groups, often leading to emotional distress or reputational damage. Unlike general hate speech, cyberbullying is typically personalized, targeted, and often context-dependent, emerging in the comments and discussions related to specific incidents. It can manifest in both direct and implicit expressions, such as sarcasm, derogatory innuendo, or culturally coded language. Annotators were trained to identify both explicit and subtle indicators of abuse, guided by generated explanations. To clarify, cyberbullying differs from general hate speech in that it typically involves personal attacks or targeted harassment rather than generic negative sentiment toward a group. For instance, a comment such as "All old people shouldn't be allowed to drive" would be categorized as hate speech, as it targets a demographic group (age) but lacks personalization. In contrast, "You're too old to drive, just stay home next time, Grandpa Li" constitutes cyberbullying, as it involves direct, targeted, and demeaning language toward a specific individual. In our annotation, only abusive content that is directed at identifiable individuals or groups in the context of a public incident is labeled as cyberbullying.

During the annotation process, we first performed a joint annotation of 1,000 samples by all annotators to establish a set of reference labels, hereafter referred to as "annotated labels". These 1,000 samples were then randomly inserted into the original dataset, without the annotators being aware of it. After completing the annotations, we compared the results of these 1,000 samples with the "annotated labels". If significant discrepancies were found, indicating that the annotations might be unreliable, we replaced them with annotations from different annotators. Therefore, this procedure also served as a screening mechanism to evaluate and refine the annotator group, and the information of the selected annotators is provided in Table \ref{tab:annotator_profile}. To ensure that each comment was accurately classified, we required at least three annotators to reach a consensus on the label. Based on different events, we constructed annotated datasets, and the detailed statistics of the final dataset are shown in Table \ref{dataset_statistics}.

\begin{figure}[t]
	\centering
	\includegraphics[scale=0.31]{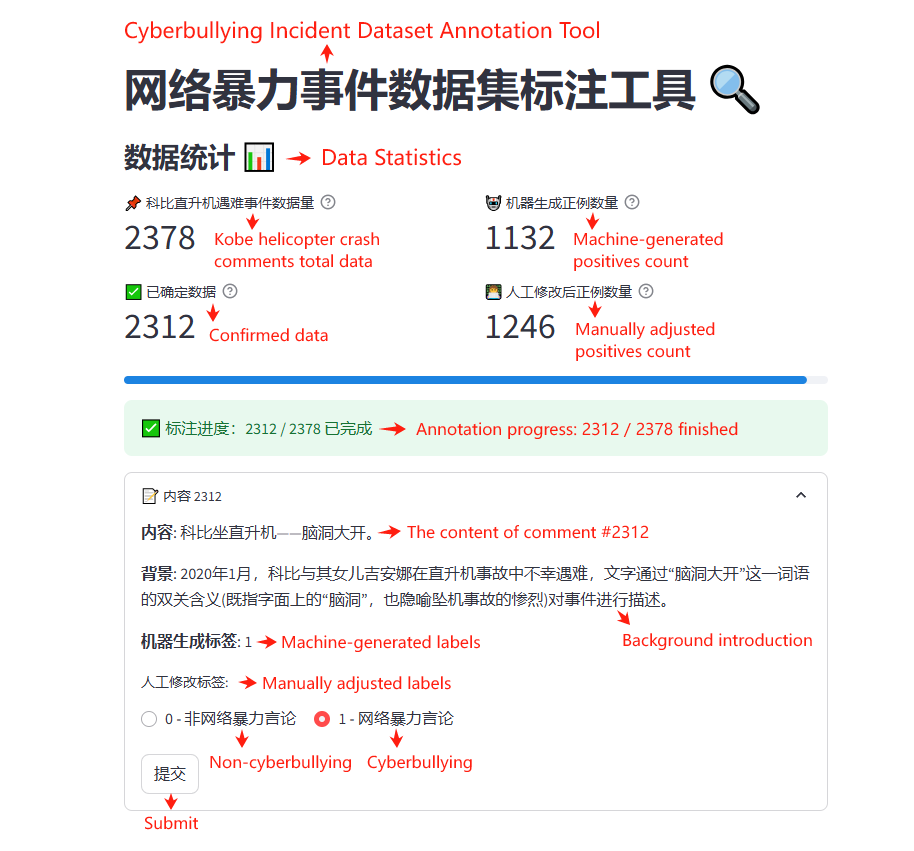}  % ???????? 80%
	\caption{
		Screenshot of an annotation example on the annotation website. The red text indicates the English translation.
	}
	\label{web}
\end{figure}

As shown in Figure \ref{web}, we have created a specialized website for annotating data. On each page of the website, a sentence is presented with the generated explanations. For each pseudo label, there are two radio buttons labeled "non-cyberbullying" and "cyberbullying". The task of annotators is to select "cyberbullying" if they considered the given sentence to be a offensive text. Conversely, they were to choose "non-cyberbullying" if they determined that the sentence would be non-offensive.

\section{Dataset Analysis}
\begin{table}[htbp]
	\centering
	\caption{Dataset Statistics Overview. Proportion: The average proportion of offensive comments within incidents.}
	\scalebox{0.8}{
		\begin{tabular}{ccc}
			\toprule
			
			\multirow{2}{*}{Total Comments (415,463)} &
			Cyberbullying Comments & 19.20\% \\
			\cmidrule(l){2-3}
			& Non-Cyberbullying Reviews & 80.80\% \\
			\midrule
			
			\multirow{2}{*}{Hotspot Events (195)} &
			Cyber Violence & 86 \\
			\cmidrule(l){2-3}
			& Normal Events & 109 \\
			\midrule
			
			\multirow{2}{*}{Average} &
			Length Of Reviews & 19.9 \\
			\cmidrule(l){2-3}
			& Reviews Per Event & 2130 \\
			\midrule
			
			\multirow{2}{*}{Proportion} &
			Cyber Violence & 25.76\% \\
			\cmidrule(l){2-3}
			& Normal Events & 8.85\% \\
			\midrule
		\end{tabular}
	}
	\label{dataset_statistics}
\end{table}
The statistical information of the constructed Chinese cyberbullying detection dataset organized by incidents, CHNCI, is presented in Table \ref{dataset_statistics}. The dataset consists of a total of 415,463 sentences with a ratio of cyberbullying instances is about 19\%. On average, each sentence contains nearly 20 words. To validate the quality of constructed dataset, we conducted the following data analysis.

\textbf{The Analysis of Dataset Construction}
We conducted a pilot test with a single annotator, who was able to annotate approximately 360 instances within one hour when using explanation-based prompts. Notably, the average time spent per task was about 10 seconds, which is quite remarkable in terms of efficiency. In contrast, the same annotator processed only about 120 instances per hour without providing explanations, indicating that explanation-guided annotation significantly improved speed. This high efficiency can be attributed to two main factors: first, native speakers are able to quickly make binary judgments regarding malicious language detection; second, annotators only need to read each target comment once to make a judgment on malicious language detection within a single task.

\textbf{The Analysis of Dataset Coverage}
We show that CHNCI achieves high coverage. The dataset consists of 195 incidents, and the proportion of cyberbullying ones is about 50\%. CHNCI covers five text genres: Business, Entertainment, Sports, Society and Politics. The "Cyber Violence" and "Normal Events" in the last two columns on Table \ref{dataset_statistics} refers to the proportion of offensive sentences in cyberbullying and non-cyberbullying incidents, respectively. The high coverage of CHNCI can be a a benchmark for evaluating incident prediction, which has shown in Section Validation.

\textbf{The Analysis of Dataset Quality} 
To validate the quality of constructed CHNCI dataset, we randomly selected 300 instances of online comments for evaluation. An annotator, proficient in Chinese and familiar with social media, was assigned to assess the accuracy of detecting malicious comments within the selected instances. The annotator carefully examined each instance to determine whether the comment was offensive or non-offensive, classifying them accordingly into cyberbullying or non-cyberbullying categories. The accuracy was calculated as 281/300, which corresponds to 93.7\%. This accuracy rate, exceeding 90\%, demonstrates the high quality of the dataset.

\textbf{The Analysis of Dataset Agreement}
To measure the consistency of annotators, we further calculate common agreement metrics such as Cohen's Kappa \cite{cohen1960coefficient} and Fleiss's Kappa \cite{fleiss1971measuring}. Cohen's Kappa measures agreement between two raters and Fleiss's Kappa is used to assess the degree of agreement among multiple raters. The Kappa result be interpreted as follows: values$ \le $0 as indicating no agreement and 0.01-0.20 as none to slight, 0.21-0.40 as fair, 0.41-0.60 as moderate, 0.61-0.80 as substantial, and 0.81-1.00 as almost perfect agreement.

The inter-annotator agreement scores for the three annotators are presented in Table \ref{dataset_evaluation}. Specifically, Fleiss' Kappa is calculated for the dataset, yielding a score of 0.609. This result indicates a substantial level of agreement among annotators, supporting the consistency and reliability of the annotation process.

\begin{table*}[htbp]
	\centering
	\renewcommand\arraystretch{1.1}
	\caption{Cohen's Kappa agreement scores for pairs of annotators}
	\setlength{\tabcolsep}{9pt}
	\resizebox{\textwidth}{!}{ % ?????????????????
		\begin{tabular}{@{}cccc@{}}
			\toprule
			Cohen's Kappa (A1-A2) & Cohen's Kappa (A1-A3) & Cohen's Kappa (A2-A3) & Fleiss's Kappa (A1-A2-A3) \\ \midrule
			0.436                 & 0.712                 & 0.690                 & 0.609                    \\ \bottomrule
		\end{tabular}
	}
	\label{dataset_evaluation}
\end{table*}
\section{Experiments}
\subsection{Experimental Setup}

\begin{figure}[htbp]
	\centering
	\includegraphics[width=0.8\linewidth]{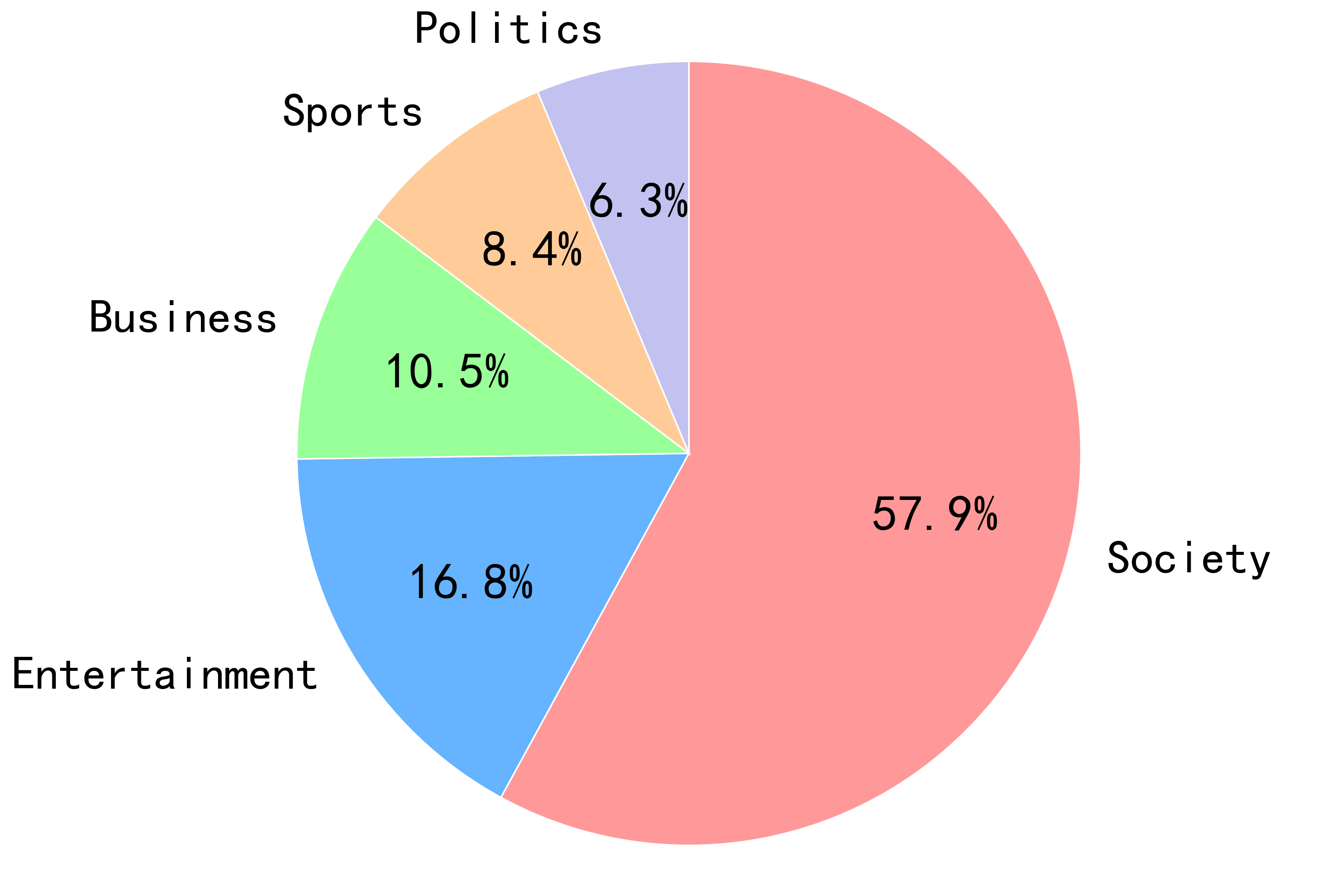}  % ? chnci_piechart.eps
	\caption{Category distribution of the CHNCI dataset.}
	\label{fig:chnci_pie}
\end{figure}

\textbf{Dataset.} We split the whole dataset CHNCI into train (80\%), valid (10\%), test (10\%) set. The experimental results are validated on test sets. The distribution of the text genres in CHNCI dataset is shown in Fig \ref{fig:chnci_pie}. Notably, Social incidents tend to provoke massive public engagement and emotional polarization, making them more prone to cyberbullying behaviors. For example, it includes events such as traffic accidents involving online influencers, student punishment incidents, metro conflicts, and train station child accidents, all of which sparked large-scale and emotionally charged discussions on social media. In contrast, discussions in categories like technology or entertainment are generally more neutral and less aggressive. Thus, the imbalance reflects the real-world distribution of online discourse, rather than a sampling bias.

\textbf{Baselines.} To evaluate the performance of different methods on our dataset, we implemented two representative baselines: HateBERT \cite{hatebert} and ConPrompt \cite{conprompt}, using their publicly available source code and default configurations as reported in the original studies. Both methods have been widely adopted in cyberbullying detection tasks and have demonstrated promising performance. The details are as follows:

\begin{itemize}
	\item \textbf{HateBERT} \cite{hatebert}: A BERT-based model pre-trained on a large corpus of banned Reddit comments, including content flagged for offensiveness, abuse, and hate speech.
	\item \textbf{ConPrompt} \cite{conprompt}: A contrastive learning-based method that generates machine-inferred statements from prompts and compares them with the original prompts, which aims to pre-train a BERT model for detecting implicit hate speech.
\end{itemize}

In addition, to investigate potential biases introduced by using LLMs for explanation generation, we conducted a comparative experiment between explanation-based annotation (involving LLM-generated reasoning) and direct labeling without explanations.

\textbf{Metrics.} We employ the designated official metrics, namely "Acc" and  "F1-score" as outlined in the SemEval 2007 task. These metrics provide a comprehensive and detailed evaluation from multiple perspectives.

\textbf{Implementation Details.} Paraphraser-based (Para) and CoT-based (CoT): detections were made directly through a conversational approach with 5-shot training data. For multi-agents-based (M-Agent) method, the number of agents is 5, and the threshold of internal and external voting are 3 and 5, respectively. It is worth mentioning that Llama 3-Chinese-8B is introduced as the LLMs for all the methods.

\subsection{Experimental Results}

The results of all methods across various metrics are summarized in Figure \ref{Experimental Results}. Each experiment was conducted at least three times, and the average results were calculated to ensure a fair comparison. The key findings are as follows:

\begin{figure*}[t]
	\centering
	\includegraphics[scale=0.5]{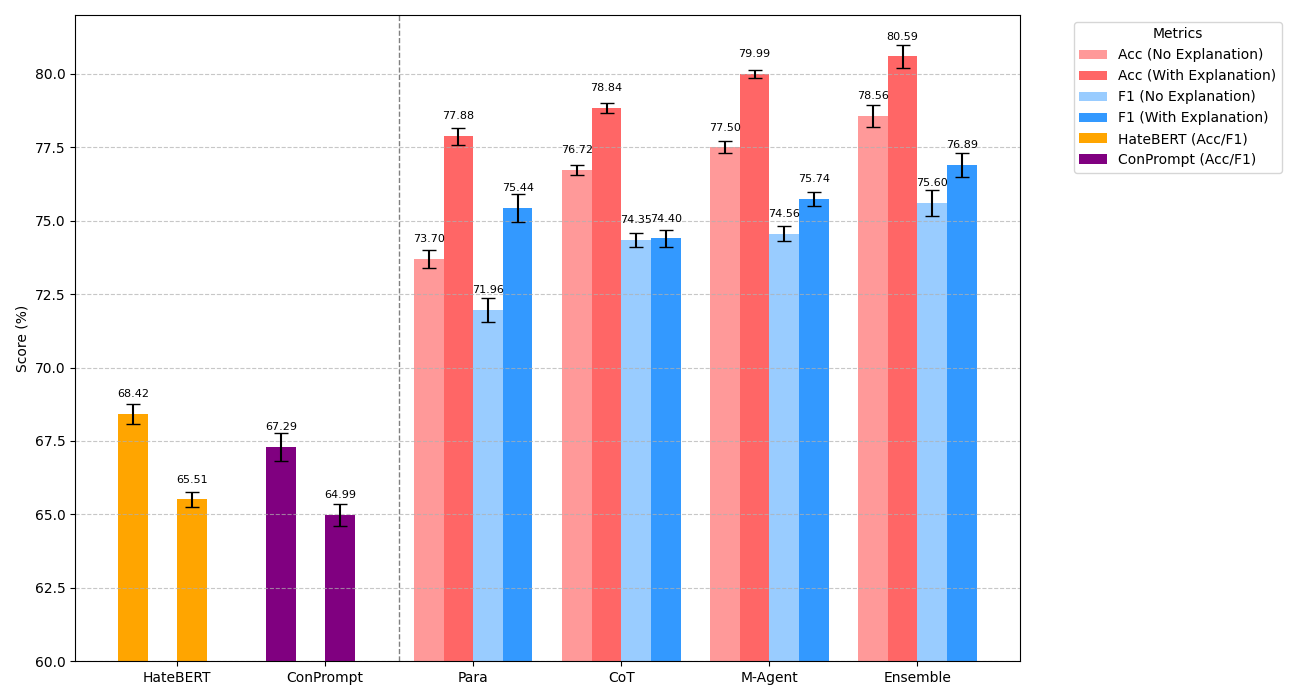}  % ? chnci_piechart.eps
	\caption{Performance Comparison with Baseline Methods}
	\label{Experimental Results}
\end{figure*}

(1) Among the individual methods, M-Agent outperforms the baselines Para and CoT. This is attributed to M-Agent's incorporation of a double-layer multi-agent voting strategy, which effectively mitigates the hallucinations of LLMs, thereby enhancing cyberbullying detection performance. Without the M-Agent strategy, CoT performs better than Para, indicating that the chain-of-thought prompting mechanism in CoT better leverages the distributed knowledge within pre-trained models. When compared to HateBERT and ConPrompt, the results demonstrate that Para, CoT, and M-Agent all achieve superior performance. This validates the effectiveness of generating explanations, as these explanations capture the nuanced humanistic histories and cultural contexts underlying posts. 

(2) The experimental results further show that the methods incorporating generated explanations significantly outperform their counterparts without explanations. The improvement can be attributed to the ability of explanations, which identify the humanistic histories and cultural stories behind the sentences, thereby capturing the true intentions for cyberbullying detection. These findings indicate that prompting models to generate explanations can indeed improve prediction accuracy. Without such explanations, even large language model (LLM)-based approaches may struggle to accurately infer the true intent behind the content.

(3) The proposed ensemble method significantly outperforms all individual models across all evaluation metrics, with statistically significant improvements. Ensemble method achieves superior results by expanding the coverage of possible explanations through the integration of multiple methods. While individual methods are often limited by specific biases and constraints, their integration within an ensemble framework mitigates these shortcomings, leading to broader coverage. Furthermore, individual cyberbullying detection methods often display varying sensitivities to different linguistic contexts, word senses, or syntactic structures. By leveraging the strengths of diverse methods, the ensemble approach demonstrates enhanced robustness in handling various linguistic scenarios, thereby contributing to its superior performance.

(4) In summary, the advantages of ensemble arise from its ability to harness the diversity, expanded coverage, and robustness of individual detection methods. These factors collectively explain the significant performance improvements of the ensemble approach over individual methods across all evaluation metrics.

\subsection{Computational Cost Analysis of the Multi-Agent Method}
Considering that the multi-agent voting strategy usually requires performing multiple rounds of inference on the same input to complete the judgment and decision fusion process, it inevitably introduces additional computational overhead. The balance among model accuracy, inference latency, and hardware resource consumption directly affects the feasibility and cost-effectiveness of deploying this method in real-world scenarios. Therefore, we conducted a comparative evaluation across different model and GPU configurations to assess the trade-off between computational cost and performance gains under practical conditions.
\begin{table}[h]
	\centering
	\caption{Performance Comparison of Different Models and GPUs. "Time (s)/per" denotes the time for one complete decision. "Price" refers to the current market price of the GPU.}
	\label{tab:model_gpu_comparison}
	\scalebox{0.9}{
	\begin{tabular}{l c c c c c}
		\hline
		\textbf{Model} & \textbf{GPU} & \textbf{ACC} & \textbf{F1} & \textbf{Time (s)/per} & \textbf{Price (\$)} \\
		\hline
		Llama3-8b      & 4090 & 80.00 & 75.94 & 3.47 & 3500 \\
		Qwen2.5-7b     & 4090 & 79.00 & 74.73 & 3.08 & 3500 \\
		Llama3-8b      & 3090 & 80.50 & 75.19 & 5.02 & 2500 \\
		Qwen2.5-7b     & 3090 & 79.32 & 75.23 & 4.65 & 2500 \\
		\hline
	\end{tabular}}
	
\end{table}

		As shown in Table \ref{tab:model_gpu_comparison}, both Llama3-8B and Qwen2.5-7B models achieved excellent performance under two GPU environments: RTX 4090 and RTX 3090. It is worth noting that the RTX 3090 not only holds a price advantage but also demonstrates comparable detection accuracy to high-end graphics cards, providing a more cost-effective option for practical deployment. These results further confirm the robustness and practicality of the multi-agent voting strategy across different hardware platforms.

\section{Validation}
\begin{figure*}[t]
	\centering
	
	\includegraphics[width=\linewidth]{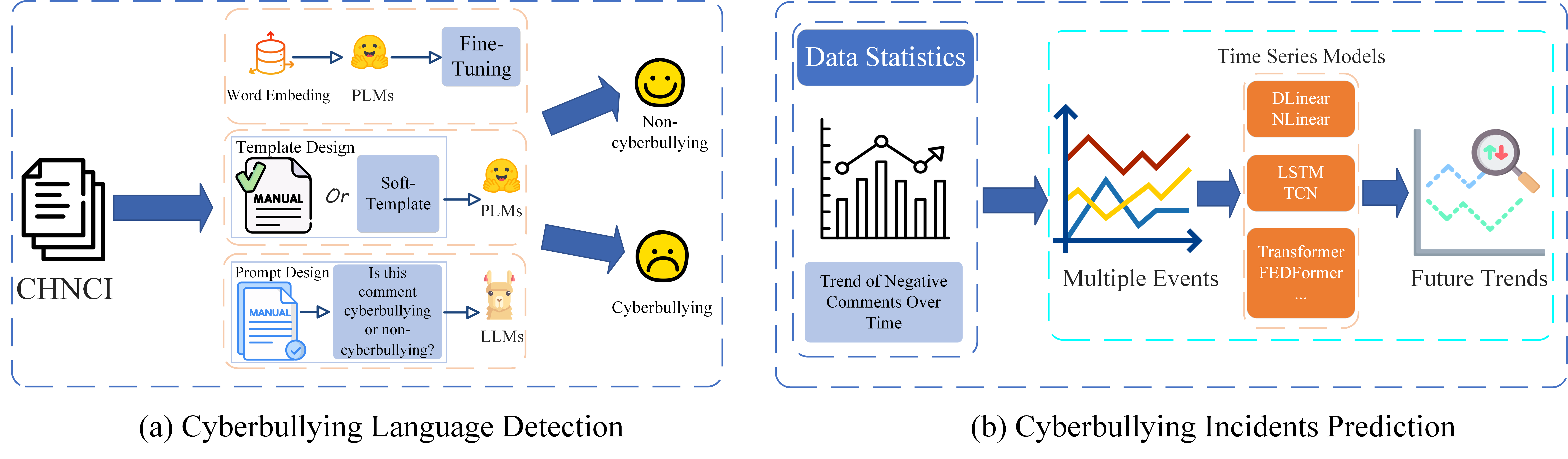}
	\caption{
		The overall framework for CHNCI validation. (a) In cyberbullying language detection, fine-tuning PLMs, prompt-based methods, and LLMs are applied for detection. (b) In cyberbullying incidents prediction, multiple time series modeling methods are utilized to forecast future trends based on historical patterns of negative comments observed across diverse events.
	}
	\label{framework2}
\end{figure*}

To validate the constructed CNHCI, we further design the validation experiments in two parts. The first part focuses on cyberbullying language detection, where various baseline methods are employed to evaluate the performance. The second part centers on time series trend forecasting, utilizing different approaches for cyberbullying incidents prediction. The process of validation is illustrated as Figure \ref{framework2}. 

\subsection{Baseline Methods for Validation}

\subsubsection{Cyberbullying Language Detection.} 

The following nine baselines are selected to validate cyberbullying language detection on CHNCI, including fine-tuning PLMs, prompt-tuning methods, and the SOTA LLMs.

\textbf{Fine-tuning PLMs:} We include BERT-base-Chinese and HateBERT as classic PLM baselines. BERT represents a widely used generic language model, while HateBERT is specifically pre-trained on offensive and hate-related content, making it a strong domain-specific baseline. The details are presented as follows.
\begin{itemize}
	\item Bert-base-Chinese \cite{bert}:A pre-trained Chinese language model specifically designed for processing tasks in Chinese. By fine-tuning the model, it can effectively detect cyberbullying comments or non-cyberbullying language.
	\item HateBERT \cite{hatebert}:A model tailored for hate speech detection, fine-tuned from the BERT framework. It excels in identifying offensive, racist, or hateful content in text.
\end{itemize}

\textbf{Prompt-tuning Methods:} We adopt ConPrompt, P-Tuning, KPT, and KPT++ to test the effectiveness of lightweight adaptation techniques in few-shot scenarios. These methods have shown promising performance in cyberbullying detection tasks, and allow models to leverage task-relevant knowledge with minimal label information. The details are presented as follows.
\begin{itemize}
	\item ConPrompt \cite{conprompt}:A contrastive learning-based prompt technique designed to enhance performance in few-shot learning tasks. It remains effective in detecting cyberbullying language even in scenarios with limited data.
	\item P-Tuing \cite{ptuning}:A parameterized prompt-based fine-tuning method that improves performance in few-shot tasks by optimizing model parameters. It is particularly advantageous in detecting cyberbullying language in imbalanced or small datasets.
	\item KPT \cite{kpt}:A method that integrates external knowledge with prompt techniques to enhance model performance by leveraging background information. It is suitable for detecting complex or implicit cyberbullying language.
	\item KPT++ \cite{kpt++}:An enhanced version of KPT that further optimizes the combination of knowledge injection and prompt learning methods. It performs exceptionally well in cross-cultural or cross-domain cyberbullying language detection.
\end{itemize}

\textbf{Large Language Models (LLMs):} We evaluate Llama3, Qwen, and DeepSeek-V3 in zero-shot settings. These open-source LLMs represent the current state-of-the-art in general-purpose language understanding and generation. Including these models allows us to assess the upper bounds of performance without task-specific fine-tuning, which is crucial for scalable real-world deployment. The details are presented as follows.
\begin{itemize}
	\item Llama3: A large-scale multilingual open-source language model developed by Meta, excelling in handling long texts and multilingual tasks. It is ideal for large-scale and diverse cyberbullying language detection.
	\item Qwen: A large-scale language model developed by Alibaba, featuring multimodal capabilities and strong language comprehension. It excels at detecting cyberbullying comments and hate speech on Chinese and multilingual platforms.
	\item DeepSeek-V3: A large language model developed by DeepSeek, supporting both Chinese and English tasks with 128K context length capability, demonstrating excellent performance in detecting cyberbullying content in complex contexts.
\end{itemize}

\subsubsection{Cyberbullying Incidents Prediction.} The following nine baselines are selected to validate cyberbullying incidents prediction on CHNCI, including the deep learning and transformer-based methods. These models are selected for their strong performance in time series forecasting and event trend modeling.
%the deep learning methods including GRU \cite{gru}, TCN \cite{tcn}, LSTM \cite{lstm}, DLinear \cite{lr}, and NLinear \cite{lr}; the transformer-based methods including Transformer \cite{transformer}, Informer \cite{informer}, Autoformer \cite{autoformer}, and FEDformer \cite{fedformer}.

\textbf{Deep Learning Methods:} The deep learning methods including GRU, TCN, and LSTM provide a strong reference point for evaluating temporal dynamics of incident-level data. DLinear and NLinear are recently proposed time series models and have shown excellent performance with lower computational cost, which help assess the efficacy of lightweight forecasting models.
\begin{itemize}
	\item GRU \cite{gru}:An optimized type of Recurrent Neural Network (RNN) that improves computational efficiency by simplifying the structure of traditional RNNs. GRU is particularly suitable for tasks involving the processing of time series data and excels in short-term time series forecasting.
	\item TCN \cite{tcn}:A time series modeling method based on Convolutional Neural Networks (CNNs), capturing long-term dependencies through causal convolution and dilated convolution. It is widely used in time series regression tasks, such as trend forecasting.
	\item LSTM \cite{lstm}:A classic type of Recurrent Neural Network that effectively addresses the problem of long-term dependencies with specially designed memory cells. It performs exceptionally well in tasks such as weather forecasting and stock trend prediction that require processing of long-term dependencies.
	\item DLinear \cite{lr}:A model that utilizes a decomposition mechanism to divide time series into trend and seasonal components, models them separately, and then merges the outputs. It demonstrates outstanding performance in time series forecasting tasks that exhibit trends and seasonal patterns.
	\item NLinear \cite{lr}:An improved version of DLinear, which enhances the model's forecasting stability and generalization ability by introducing a normalization mechanism. It is suitable for dealing with non-stationary time series scenarios.
\end{itemize}

\textbf{Transformer-based Methods:} Four Transformer-based methods including Transformer, Informer, Autoformer, and FEDformer incorporate innovations like sparse attention, decomposition mechanisms, and frequency-domain analysis. These methods are particularly suited to handling complex and long-term temporal patterns, making them ideal for incident-level forecasting.
\begin{itemize}
	\item Transformer \cite{transformer}:A deep learning model based on attention mechanisms that has gained widespread attention due to its powerful global context modeling capabilities. It is suitable for complex tasks such as multi-variable time series forecasting and financial analysis.
	\item Informer \cite{informer}:An optimized variant of Transformer for long-term time series forecasting, significantly improving computational efficiency through sparse attention mechanisms. It is particularly suitable for time series forecasting tasks with long time spans.
	\item Autoformer \cite{autoformer}:A variant of Transformer that focuses on time series trend modeling and seasonal pattern capturing. Its automated design reduces the need for parameter tuning and performs particularly well in tasks with strong trends and cycles.
	\item FEDformer \cite{fedformer}:A variant of Transformer that combines Fourier transformation and decomposition techniques, enhancing the model's performance by strengthening frequency domain analysis. It is suitable for complex time series forecasting tasks, such as traffic flow and meteorological data analysis.
\end{itemize}

\subsection{Implementation Details and Evaluation Metrics}

\textbf{Criteria for validating cyberbullying incident.} Based on the data statistics of CHNCI, we propose the evaluation criteria for validating whether it constitutes a cyberbullying incident, including two rules: (1) Offensive Comments Peak Phenomenon: If the number of offensive comments within a certain time interval exceeds 5\% of the current total number of comments, it is considered that this may constitute a cyberbullying incident. (2) Multiple Clusters of Offensive Sentiments: When the proportion of offensive comments in multiple time intervals (threshold is set to 5 in the experiments) exceeds 50\% of that time interval, it is considered as a cyberbullying incident.

\textbf{Experimental Setup.} In the task of cyberbullying language detection, we adopted different training strategies for various model types. When fine-tuning pre-trained language models (PLMs), we randomly selected training sets consisting of 600, 1,000, and 1,600 samples, with the remaining data reserved for testing. For prompt-tuning-based methods, we simulated few-shot learning scenarios by employing 30-shot, 40-shot, and 50-shot settings, using the rest of the samples for evaluation. In contrast, large language models (LLMs) were evaluated in a zero-shot setting without any task-specific fine-tuning. LLMs were evaluated under both zero-shot and 5-shot settings. Specifically, Llama3 and Qwen models were deployed and executed locally, while DeepseekV3 was accessed via API-based inference.

In the task of cyberbullying incidents prediction, we first calculated the number of offensive comments for each event on an hourly basis, constructing complete time series data for each event (i.e., the number of offensive comments per hour). We then selected data from five representative events as the training set, including three cyberbullying incidents and two normal events, while using data from the remaining events as the test set. We employed a sliding window method, using the number of offensive comments from the past five time intervals as input to predict the number of offensive comments for the next time interval.

\textbf{Parameter settings}
In the cyberbullying language detection, to ensure fairness, we set the following parameters for methods of the same type: for prompt-tuning methods (P-tuing, KPT, KPT++), the learning rate is set to 4e-5, batch size to 32, and the number of training epochs to 20. For fine-tuning PLMs (BERT, HateBERT, and Conprompt), the learning rate is set to 2e-5, batch size to 16, and the number of training epochs to 20. 

\begin{table}
	\centering
	\caption{Results (\%) of cyberbullying language detection. The best values are bolded.}
	\label{main_exp1}
	\small
	\resizebox{0.5\textwidth}{!}{% 宽度为页面宽度，高度按比例缩放
		\begin{tabular}{ccccccc} 
			\toprule
			\multirow{2}{*}{Method} & \multicolumn{2}{c}{600/30}                                                                                      & \multicolumn{2}{c}{1000/40}                                                                                                       & \multicolumn{2}{c}{1600/50}                                                                                                        \\ 
			\cline{2-7}
			& Acc                                                    & F1s                                                    & Acc                                                    & F1s                                                                      & Acc                                                                      & F1s                                                     \\ 
			\midrule
			Bert                    & \begin{tabular}[c]{@{}c@{}}63.00\\(±0.14)\end{tabular} & \begin{tabular}[c]{@{}c@{}}66.69\\(±0.36)\end{tabular} & \begin{tabular}[c]{@{}c@{}}67.14\\(±0.10)\end{tabular} & \begin{tabular}[c]{@{}c@{}}69.86\\(±0.20)\end{tabular}                   & \begin{tabular}[c]{@{}c@{}}70.10\\(±0.08)\end{tabular}                   & \begin{tabular}[c]{@{}c@{}}72.48\\(±0.17)\end{tabular}  \\
			HateBert                & \begin{tabular}[c]{@{}c@{}}62.59\\(±0.07)\end{tabular} & \begin{tabular}[c]{@{}c@{}}66.56\\(±0.18)\end{tabular} & \begin{tabular}[c]{@{}c@{}}68.22\\(±0.31)\end{tabular} & \begin{tabular}[c]{@{}c@{}}70.29\\(±0.51)\end{tabular}                   & \begin{tabular}[c]{@{}c@{}}70.49\\(±0.31)\end{tabular}                   & \begin{tabular}[c]{@{}c@{}}71.70\\(±0.52)\end{tabular}  \\
			Conprompt               & \begin{tabular}[c]{@{}c@{}}64.97\\(±0.12)\end{tabular} & \begin{tabular}[c]{@{}c@{}}68.20\\(±0.21)\end{tabular} & \begin{tabular}[c]{@{}c@{}}72.33\\(±0.44)\end{tabular} & \begin{tabular}[c]{@{}c@{}}72.23\\(±0.65)\end{tabular}                   & \begin{tabular}[c]{@{}c@{}}73.89\\(±0.45)\end{tabular}                   & \begin{tabular}[c]{@{}c@{}}73.86\\(±0.63)\end{tabular}  \\
			P-tuning                & \begin{tabular}[c]{@{}c@{}}62.90\\(±0.79)\end{tabular} & \begin{tabular}[c]{@{}c@{}}66.83\\(±0.70)\end{tabular} & \begin{tabular}[c]{@{}c@{}}68.66\\(±1.30)\end{tabular} & \begin{tabular}[c]{@{}c@{}}71.49\\(±0.94)\end{tabular}                   & \begin{tabular}[c]{@{}c@{}}71.36\\(±1.21)\end{tabular}                   & \begin{tabular}[c]{@{}c@{}}73.61\\(±0.95)\end{tabular}  \\
			KPT                     & \begin{tabular}[c]{@{}c@{}}61.61\\(±1.14)\end{tabular} & \begin{tabular}[c]{@{}c@{}}65.65\\(±0.99)\end{tabular} & \begin{tabular}[c]{@{}c@{}}70.70\\(±0.03)\end{tabular} & \begin{tabular}[c]{@{}c@{}}73.18\\(±0.02)\end{tabular}                   & \begin{tabular}[c]{@{}c@{}}73.97\\(±1.71)\end{tabular}                   & \begin{tabular}[c]{@{}c@{}}75.63\\(±0.79)\end{tabular}  \\
			KPT++                   & \begin{tabular}[c]{@{}c@{}}69.60\\(±1.00)\end{tabular} & \begin{tabular}[c]{@{}c@{}}72.22\\(±0.71)\end{tabular} & \begin{tabular}[c]{@{}c@{}}70.79\\(±0.47)\end{tabular} & \begin{tabular}[c]{@{}c@{}}73.29\\(±0.24)\end{tabular}                   & \begin{tabular}[c]{@{}c@{}}72.63\\(±2.16)\end{tabular}                   & \begin{tabular}[c]{@{}c@{}}74.56\\(±1.59)\end{tabular}  \\ 
			\hline
			\multirow{2}{*}{LLM}    & \multicolumn{3}{c}{zero-shot}                                                                                                                                            & \multicolumn{3}{c}{5-shot}                                                                                                                                                                                    \\ 
			\cline{2-7}
			& Acc                                                    & F1s                                                    & Mode                                                   & Acc                                                                      & F1s                                                                      & Mode                                                    \\ 
			\hline
			
			Llama3-8B               & \begin{tabular}[c]{@{}c@{}}73.04\\(±0.09)\end{tabular} & \begin{tabular}[c]{@{}c@{}}73.96\\(±0.09)\end{tabular} & LOCAL                                                  & \begin{tabular}[c]{@{}c@{}}75.00\\(±0.23)\end{tabular}                   & \begin{tabular}[c]{@{}c@{}}76.96\\(±0.17)\end{tabular}                   & LOCAL                                                   \\
			Qwen-7B                 & \begin{tabular}[c]{@{}c@{}}75.86\\(±0.09)\end{tabular} & \begin{tabular}[c]{@{}c@{}}77.74\\(±0.02)\end{tabular} & LOCAL                                                  & \begin{tabular}[c]{@{}c@{}}78.50\\(±0.22)\end{tabular}                   & \begin{tabular}[c]{@{}c@{}}77.70\\(±0.31)\end{tabular}                   & LOCAL                                                   \\
			\textbf{DeepseekV3}              & \begin{tabular}[c]{@{}c@{}}73.43\\(±0.06)\end{tabular} & \begin{tabular}[c]{@{}c@{}}76.09\\(±0.08)\end{tabular} & API                                                    & \begin{tabular}[c]{@{}c@{}}\textbf{81.50}\\\textbf{(±0.28)}\end{tabular} & \begin{tabular}[c]{@{}c@{}}\textbf{79.78}\\\textbf{(±0.16)}\end{tabular} & \textbf{API}                                            \\
			\bottomrule
	\end{tabular}}
\end{table}

Additionally, in the cyberbullying incidents prediction, to ensure fairness, we set the learning rate to 1e-3, the number of training epochs for each event to 10, and the sliding window size to 5 for all the methods.

\textbf{Evaluation metrics}
In the cyberbullying language detection, the same metrics, Acc and F1-socre, are used as the evaluation metrics. In the cyberbullying incidents prediction, MAE (Mean Absolute Error) and RMSE (Root Mean Squared Error) are used as evaluation metrics. Notably, the smaller value of MAE and RMSE indicates better results.

\subsection{Validation Results}

The results of the cyberbullying language detection and cyberbullying incidents prediction on CHNCI are shown in Table \ref{main_exp1} and Table \ref{main_exp2}, respectively. It is worth noting that, each experiment was conducted three times, and the average and standard deviation were computed.

\textbf{Cyberbullying Language Detection.}
In PLMs, the data reveals that HateBERT achieved 62.59\% (±0.07) accuracy with 600 samples, slightly lower than standard BERT's 63.00\% (±0.14). However, it demonstrated better scalability at 1000 and 1600 sample sizes, reaching 68.22\% (±0.31) and 70.49\% (±0.31) accuracy respectively. This suggests that the advantages of domain-specific pretraining become more apparent as data volume increases. Conprompt showed more outstanding performance, achieving 64.97\% (±0.12), 72.33\% (±0.44), and 73.89\% (±0.45) accuracy across the three data scales, significantly outperforming the baseline BERT model.

In few-shot learning scenarios, KPT++ demonstrated outstanding performance. With only 30 samples, it achieved an accuracy of 69.60\% (±1.00) and an F1-score of 72.22\% (±0.71), significantly outperforming both P-tuning and the standard KPT method. In particular, KPT++ improved the F1-score by approximately 6.5 percentage points compared to KPT, fully validating its strong generalization ability and robustness under limited data conditions.

%The performance of large language models (LLMs) is particularly noteworthy. Qwen-7B achieved comprehensive leadership with 75.86\% (±0.09) accuracy and 77.74\% (±0.02) F1-score, maintaining its performance advantage consistently across all data scales. Llama3-8B and DeepseekV3 attained 73.04\% (±0.09) and 73.43\% (±0.06) accuracy respectively. While not matching Qwen-7B, they still outperformed most specialized models and prompt-tuning approaches. This phenomenon strongly demonstrates the powerful generalization capabilities that LLMs acquire through massive pretraining, enabling them to maintain stable high performance across different data scales.
%
%As training data size increases, the performance of all PLMs and prompt-tuning methods improves significantly. This indicates that data quantity plays a crucial role in determining model performance in cyberbullying language detection tasks. However, prompt-tuning methods (especially KPT) show more pronounced growth on small datasets, demonstrating their adaptability in few-shot scenarios. LLMs (such as Llama3-8B, Qwen-7B, and DeepseekV3) already achieve high performance under zero-shot conditions, with Qwen-7B even surpassing all prompt-tuning methods. This suggested that LLMs, through extensive pre-training corpora and scale effects, can achieve outstanding performance even in the absence of task-specific data, making them well-suited for unsupervised or zero-shot applications.

		Large Language Models (LLMs) exhibit particularly notable performance in this task. Under the zero-shot setting, Qwen-7B achieves the best overall results with an accuracy of 75.86\% (±0.09) and an F1 score of 77.74\% (±0.02). Llama3-8B and DeepseekV3 reach accuracies of 73.04\% (±0.09) and 73.43\% (±0.06), respectively. Although slightly inferior to Qwen-7B, they still outperform most dedicated models and prompt-tuning methods. It is worth noting that when a small amount of supervision is introduced (i.e., the 5-shot setting), all models exhibit substantial performance improvements. DeepseekV3 performs particularly well in this case, achieving an accuracy of 81.50\% (±1.08) and an F1 score of 79.78\% (±0.96). This noticeable performance leap from zero-shot to few-shot learning clearly demonstrates the strong sample efficiency of LLMs in cyberbullying detection.
		
		Moreover, as the size of the training data increases, the performance of all methods improves significantly, indicating that data volume remains a decisive factor in cyberbullying language detection. However, compared to direct fine-tuning approaches, prompt-tuning methods (especially KPT) show more pronounced gains in low-resource settings, highlighting their adaptability under few-shot conditions. Additionally, DeepseekV3 outperform locally deployed models in the 5-shot setup, suggesting that larger and more powerful model architectures can better leverage limited annotated samples. Overall, these observations emphasize the robust few-shot learning capabilities conferred by large-scale pretraining, enabling LLMs to rapidly adapt to new tasks and achieve strong performance even with only minimal task-specific supervision.

\begin{table}[htbp]
	\centering
	\normalsize % 设置整个表格的字体为normalsize
	\renewcommand{\arraystretch}{1.3} % 调整行高
	\caption{Results of cyberbullying incidents prediction. The best values are bolded.} % 添加标题
	\scalebox{0.98}{
		\begin{tabular}{@{}lcccc@{}}
			\toprule
			\multicolumn{1}{c}{Method} & \multicolumn{2}{c}{MAE}    & \multicolumn{2}{c}{RMSE}   \\ \midrule
			GRU                        & 1.4407 (\(\pm\) 0.0611) &  & 2.8429 (\(\pm\) 0.1434) &  \\ 
			TCN                        & 1.4169 (\(\pm\) 0.0649) &  & 2.7558 (\(\pm\) 0.0695) &  \\ 
			LSTM                       & 1.4039 (\(\pm\) 0.0468) &  & 2.6394 (\(\pm\) 0.1155) &  \\ 
			DLinear                    & 1.2746 (\(\pm\) 0.0096) &  & 2.6000 (\(\pm\) 0.0194) &  \\ 
			NLinear                    & 1.2512 (\(\pm\) 0.0181) &  & 2.5527 (\(\pm\) 0.0517) &  \\ 
			Transformer                & 1.3952 (\(\pm\) 0.0732) &  & 2.7860 (\(\pm\) 0.0703) &  \\ 
			Informer                   & 1.3996 (\(\pm\) 0.0847) &  & 2.7931 (\(\pm\) 0.1533) &  \\ 
			Autoformer                 & 1.4463 (\(\pm\) 0.1002) &  & 2.9284 (\(\pm\) 0.2083) &  \\ 
			FEDformer                  & \textbf{1.2077 (\(\pm\) 0.0303)} &  & \textbf{2.5259 (\(\pm\) 0.0166)} &  \\ 
			\bottomrule
		\end{tabular}
	}
	\label{main_exp2}
\end{table}

\textbf{Cyberbullying Incidents Prediction.}
FEDformer outperforms all other methods, significantly excelling in terms of performance. Its adaptive feature selection and dynamic modeling capabilities allow it to capture essential features in time series data, making it especially suitable for tasks involving long time series predictions or tasks with long-term dependencies. FEDformer is ideal for scenarios that require high-precision forecasting, especially in the context of complex and dynamic time series data. NLinear follows closely behind, demonstrating strong capabilities in modeling non-linear relationships. This method is well-suited for handling time series data with significant nonlinear characteristics, which performs excellently in scenarios where data exhibits complex trends or fluctuations. DLinear also shows stable performance, with its advantage lying in the simplified linear model that provides efficient and robust predictions for simpler linear time series tasks. DLinear is suitable for scenarios with limited computational resources or where data variation is relatively small.

LSTM and GRU provide stable results, but their performance in more complex tasks is inferior to newer methods. These traditional architectures can capture long-term dependencies in time series data, but their adaptability to dynamic changes is limited, making them more appropriate for scenarios with smaller datasets and relatively stable variations. Transformer and Autoformer, despite their powerful global dependency modeling capabilities, show weaker performance in the experiments, possibly due to poor adaptability of the model architecture to the dataset with overfitting. Transformer is more suitable for large-scale data scenarios where global information capture is essential, while Autoformer is better suited for data with strong seasonal components.

\begin{figure*}[htbp]
	\centering
	\begin{minipage}{0.5\textwidth}
		\centering
		\includegraphics[width=\linewidth]{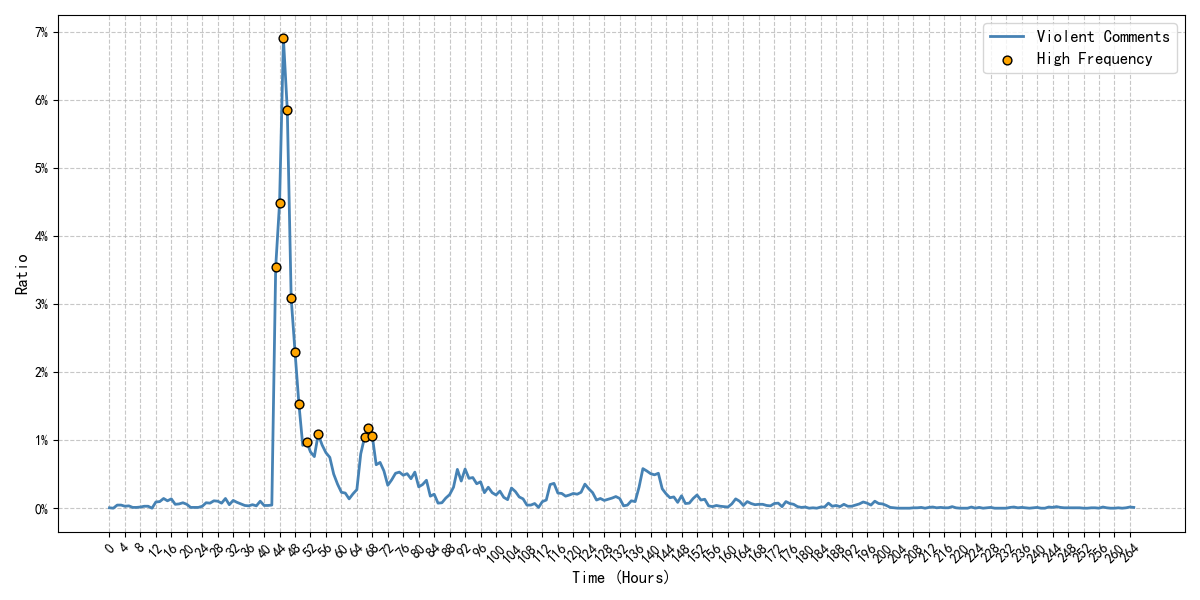}
		\textbf{(a) Cyberbullying Incidents}
	\end{minipage}%
	\hfill
	\begin{minipage}{0.5\textwidth}
		\centering
		\includegraphics[width=\linewidth]{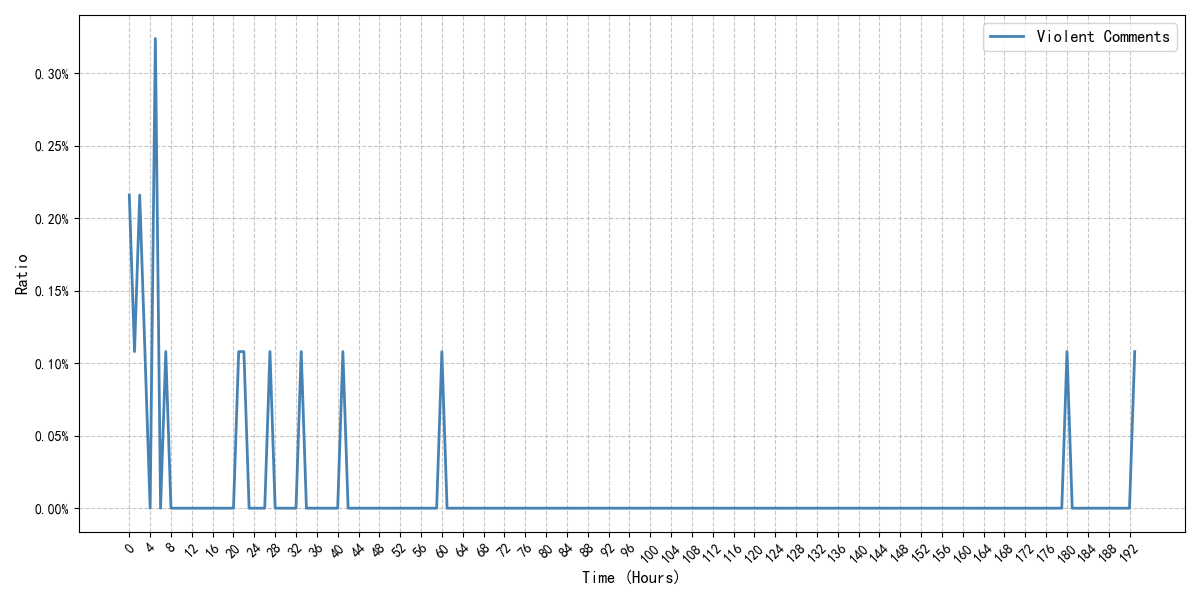}
		\textbf{(b) Non-Cyberbullying Incidents}
	\end{minipage}
	\caption{Hourly Trend of Comments: Comparison of Cyberbullying Incidents and Normal Events. The x-axis represents the hours elapsed since the event began, while the y-axis represents the ratio of offensive comments within each hour.} % 将标题放在 \caption 的位置，LaTeX 会自动跨栏居中
	\label{graph1}
\end{figure*}

\begin{figure*}[t]
	\centering
	\begin{minipage}{0.5\textwidth}
		\centering
		\includegraphics[width=\linewidth]{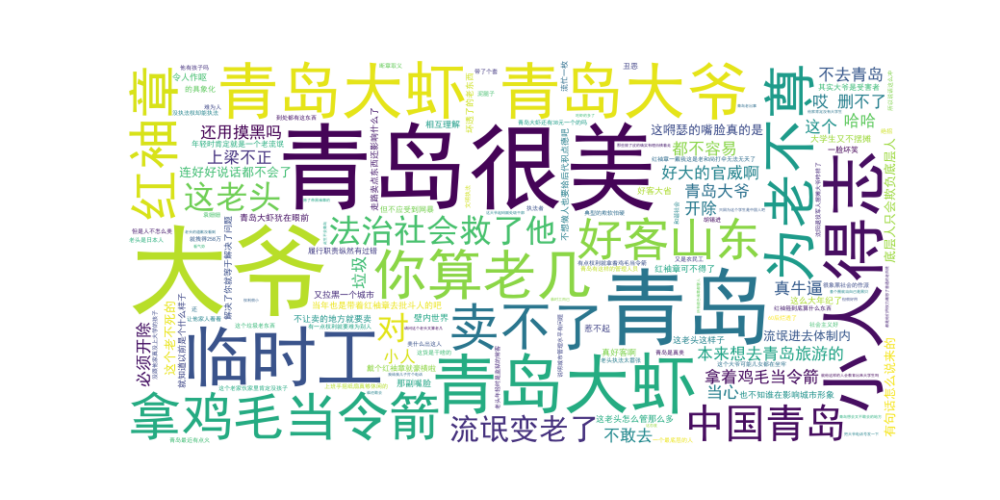}
		\textbf{(a) Cyberbullying Incidents}
	\end{minipage}%
	\hfill
	\begin{minipage}{0.5\textwidth}
		\centering
		\includegraphics[width=\linewidth]{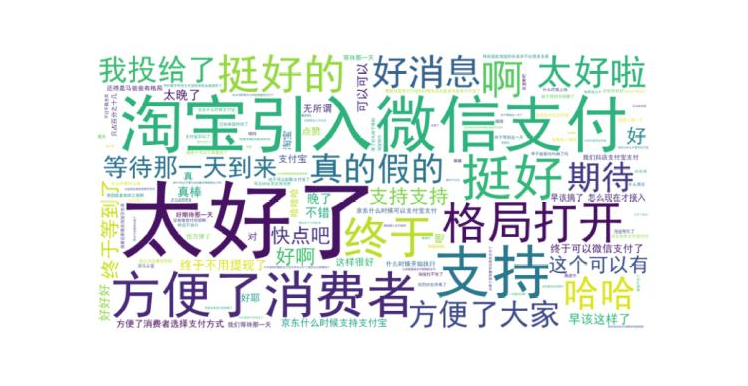}
		\textbf{(b) Non-Cyberbullying Incidents}
	\end{minipage}
	\caption{Word clouds of online comments during the event. (a) shows cyberbullying content, and (b) shows non-cyberbullying content. The figure highlights differences in high-frequency words between the two.} % 将标题放在 \caption 的位置，LaTeX 会自动跨栏居中
	\label{graph2}
\end{figure*}

To visually express the differences between Cyberbullying and Non-Cyberbullying Incidents, we present two incidents in CHNCI,along with their corresponding word clouds, as illustrated in Figure \ref{graph1} and Figure \ref{graph2}. 


For Figure \ref{graph1}, the y-axis represents the ratio of cyberbullying comments within each hour. We can see that in Figure \ref{graph1}(a), there is a clear peak in the number of cyberbullying comments (the peak point is about 7\%), which account for a large proportion of the total number of comments. In contrast, the number of cyberbullying comments in Figure \ref{graph1}(b) is significantly more evenly distributed (the peak point is just 0.35\%), and accounts for a smaller proportion of the total number of comments. It is evident that these figures are consistent with our cyberbullying incidents prediction.

Figure \ref{graph2} illustrates the differences in word clouds between cyberbullying and non-cyberbullying incidents. In Figure \ref{graph2}(a), high-frequency words such as "Qingdao prawns," "Who do you think you are," and "temporary worker" carry strong sarcastic and aggressive tones, reflecting the negative emotions and accusatory attitudes often expressed during cyberbullying events. In contrast, Figure \ref{graph2}(b) features more positive words like "Great," "Support," and "WeChat Pay," indicating approval and anticipation from users. This contrast in word frequency clearly highlights the emotional divergence in public opinion between the two types of incidents.

\section{Conclusions}

This study presents the first comprehensive exploration of the Chinese cyberbullying detection task. We propose a novel annotation method to construct a large-scale Chinese dataset organized by incidents through a collaborative human-machine approach. The constructed dataset consists of 415,463 instances in 195 incidents, with a ratio of cyberbullying instances is about 19\%. Our proposed ensemble method by leveraging the strengths of each method while mitigating their weaknesses, our ensemble approach significantly outperforms the individual cyberbullying detection methods with generated explanations across all evaluation metrics. The constructed dataset can not only be a benchmark for evaluating cyberbullying detection, but also more important, for the task of cyberbullying incident prediction.

In the future, we plan to extend our work in two directions. Firstly, we aim to explore automatic criteria for validating cyberbullying incident, feature representation learning methods can be introduced to identify cyberbullying incidents. Secondly, the constructed CHNCI dataset will be used as a benchmark for future research.

In conclusion, our study fills the research gap on how to construct a large-scale cyberbullying detection dataset with high coverage and low cost, providing a solid foundation for further development. The construction of a high-quality dataset and the development of an effective ensemble method showcase the potential for improved cyberbullying detection in the Chinese language.

\section*{Acknowledgments}

This research is partially supported by the National Natural Science Foundation of China under grants (62076217), National Language Commission of China under grants (ZDI145-71), Key Research and Development Program of Jiangsu Province in China (BE2023315), Yangzhou Science and Technology Plan Project City School Cooperation Special Project (YZ2023199), Open Project Program of Key Laboratory of Knowledge Engineering with BigData (the Ministry of Education of China, NO. BigKEOpen2025-06).

\bibliographystyle{named}
\bibliography{ijcai23}

\end{CJK}
\end{document}